\documentclass[a4paper,conference]{IEEEtran}

\usepackage{algorithm}
\usepackage{algorithmic}
\usepackage{graphicx}

\usepackage{amsmath,amsfonts,bm}

\def\eqref#1{equation~\ref{#1}}

\def\1{\bm{1}}

\DeclareMathAlphabet{\mathsfit}{\encodingdefault}{\sfdefault}{m}{sl}
\SetMathAlphabet{\mathsfit}{bold}{\encodingdefault}{\sfdefault}{bx}{n}

\usepackage{subcaption}     %
\usepackage{xspace}         %
\usepackage{siunitx}       %
\usepackage{booktabs}
\usepackage{epsfig}
\usepackage{amsmath}
\usepackage{amssymb}
\usepackage{multirow}
\usepackage{comment}
\usepackage{amsfonts}       %
\usepackage{nicefrac}       %
\usepackage{microtype}      %
\usepackage{xcolor}         %
\usepackage[normalem]{ulem}
\usepackage{wrapfig}
\usepackage{cite}
\usepackage[pagebackref=true,breaklinks=true,letterpaper=true,colorlinks,bookmarks=false]{hyperref}

\usepackage{newfloat}
\usepackage{listings}
\floatstyle{ruled}
\newfloat{listing}{tb}{lst}{}
\floatname{listing}{Listing}

\newcommand{\ourmodel}{CVG\xspace}
\newcommand{\dvdgan}{DVD-GAN\xspace}

\title{Cascaded Video Generation for Videos In-the-Wild}

\author{\IEEEauthorblockN{Lluis Castrejon}
\IEEEauthorblockA{Université de Montréal - Mila Quebec \\
Email: lluis.castrejon@gmail.com}
\and
\IEEEauthorblockN{Nicolas Ballas} 
\IEEEauthorblockA{Facebook AI Research \\ 
Email: ballasn@fb.com}
\and
\IEEEauthorblockN{Aaron Courville} 
\IEEEauthorblockA{Université de Montréal - Mila Quebec - CIFAR AI Chair \\ 
Email: aaron.courville@umontreal.ca}}

\begin{document}

\maketitle

\begin{abstract}

Videos can be created by first outlining a global view of the scene and then adding local details.
Inspired by this idea we propose a cascaded model for video generation which follows a coarse to fine approach. 
First our model generates a low resolution video, establishing the global scene structure, which is then refined by subsequent cascade levels operating at larger resolutions.
We train each cascade level sequentially on partial views of the videos, which reduces the computational complexity of our model and makes it scalable to high-resolution videos with many frames. 
We empirically validate our approach on UCF101 and Kinetics-600, for which our model is competitive with the state-of-the-art. 
We further demonstrate the scaling capabilities of our model and train a three-level model on the BDD100K dataset which generates 256x256 pixels videos with 48 frames.

\end{abstract}

\section{Introduction}

Humans have the ability to simulate visual objects and their dynamics using their imagination. 
This ability is linked to the ability to perform temporal planning or counter-factual thinking.  
Replicating this ability in machines is a longstanding challenge that generative models try to address.
Advances in generative modeling and increased computational resources have enabled the generation of realistic high-resolution images~\cite{brock2018large} or coherent text in documents~\cite{brown2020language}. 
Yet, video generation models have been less successful, in part due to their high memory requirements that scale with the generation resolution and length.

\begin{figure}[t]
    \centering
    \includegraphics[trim=20 10 20 10, width=\linewidth, clip]{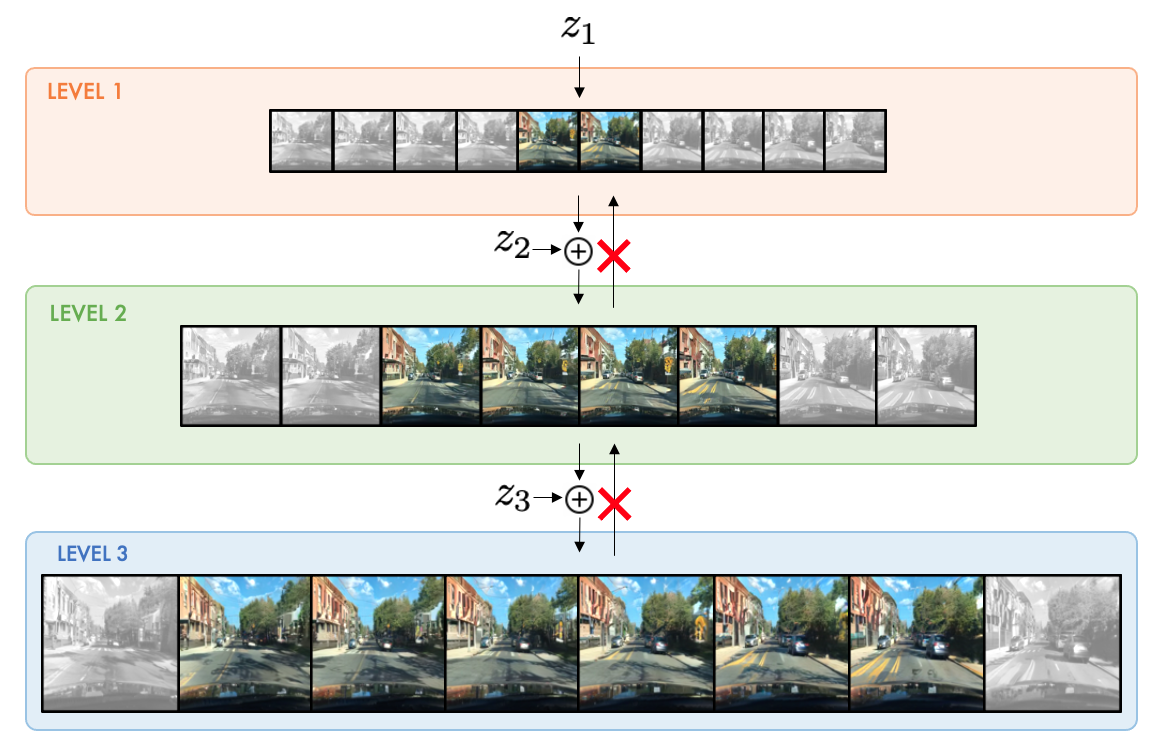}
    \caption{\small \textbf{Cascaded Video Generation}
        We propose to divide the generative process into multiple simpler problems.
        \ourmodel first generates a low resolution video that depicts a full scene at a reduced framerate. 
        This scene outline is then progressively upscaled and temporally interpolated.
        Levels are trained sequentially and do not backprogagate gradient to previous levels. Additionally, upscaling levels can be trained on temporal crops of previous level outputs (illustrated by the non-shaded images) to reduce their computational requirements. 
        Our model outperforms or matches the state-of-the-art in video generation and enables the generation of longer high resolution videos due to better scaling properties than previous methods.
    }
    \label{fig:teaser}
\end{figure}

When creating visual data, artists often first produce a rough outline of the scene, to which then they add local details in multiple iterations~\cite{locher2010does}.
The outline ensures global scene consistency and divides the creative process into multiple tractable local steps.
Inspired by this process we propose \ourmodel, a cascaded video generation model which divides the generative process into a set of simpler problems.
\ourmodel first generates a rough video that depicts a scene at a reduced framerate and resolution. 
This scene outline is then progressively upscaled and temporally interpolated to obtain the desired final video by one or more upscaling levels as depicted in Figure~\ref{fig:teaser}.
Every cascade level outputs a video that serves as the input to the next one, with each level specializing in a particular aspect of the generation.

Levels in our model are trained greedily, i.e.\ in sequence and not end-to-end. 
This allows to train only one level at a time and thus reduce the overall training memory requirements.
We formulate each level as a adversarial game that we solve leveraging the GAN framework. Our training setup has the same global solution as an end-to-end model.

To further reduce the computational needs of our model, upscaling cascade levels can be applied only on temporal crops from previous outputs during training.
Despite this temporally-local training, upscaling levels are capable of producing videos with temporal coherence at inference time, as they upscale the output of the first level, which is temporally complete albeit at a reduced resolution.
This makes \ourmodel more scalable than previous methods, making it capable of generating high resolution videos with a larger number of frames.

Our contributions can be summarized as follows:
\begin{itemize}
\item We define a cascade model for video generation which divides the generation process into multiple tractable steps. 
\item We empirically validate our approach on UCF101, Kinetics-600 and BDD100K, large-scale datasets with complex videos in real-world scenarios. \ourmodel matches or outperforms the state-of-art video generation models on these datasets.
\item We demonstrate that our approach has better scaling properties than comparable non-cascaded approaches and train a three-level model to generate videos with 48 frames at a resolution of 256x256 pixels. 
\end{itemize}
\section{Cascaded Video Generation}
\label{section:method}

Video scenes can be created by first outlining the global scene and then adding local details.
Following this intuition we propose \ourmodel, a cascade model in which each level only treats a lower dimensional view of the data.
Video generation models struggle to scale to high frame resolutions and long temporal ranges.
The goal of our method is to break down the generation process into smaller steps which require less computational resources when considered independently.

\textbf{Problem Setting}
We consider a dataset of videos $(\mathbf{x}_1, ..., \mathbf{x}_n)$ where each video $\mathbf{x}_i = (\mathbf{x}_{i; 0}, ... , \mathbf{x}_{i; T})$ is a sequence of $T$ frames $\mathbf{x}_{i; t} \in \mathbb{R}^{H\times W\times 3}$.
Let $f_s$ denote a spatial bilinear downsampling operator and  $f_t$ a temporal subsampling operator. 
For each video $\mathbf{x}_i$, we can obtain lower resolution views of our video by repeated application of $f_s$ and $f_t$, i.e. $\mathbf{x}^l_{i} = f_s(f_t(\mathbf{x}^{l+1}_{i}), \forall l \in [1..L]$ with  $\mathbf{x}^L_{i}=\mathbf{x}_{i}$. 

Each ($\mathbf{x}^1_{i}$, ..., $\mathbf{x}^L_{i}$) comes from a joint data distribution $p_{d}$. 
The task of video generation consists in learning a generative distribution $p_g$ such that $p_g=p_{d}$.

\textbf{Cascaded Generative Model}
We define a generative model that approximates the joint data distribution according to the following factorization:
\begin{eqnarray}
p_{g}(\mathbf{x}^1, ..., \mathbf{x}^L) =  p_{g_{L}}(\mathbf{x}^L | \mathbf{x}^{L-1}) ... p_{g_2}(\mathbf{x}^2 | \mathbf{x}^1) p_{g_1}(\mathbf{x}^1).
\label{eq:gen_f}
\end{eqnarray}
Each $p_{g_i}$ defines a level in our model. 
This formulation allows us to decompose the generative process in to a set of smaller problems.
The first level $p_{g_1}$ produces low resolution and temporally subsampled videos from a latent variable. 
For subsampling factors $K_{T}$ and $K_{S}$ (for time and space respectively), the initial level generates videos $\mathbf{x}^{1}_i = (\mathbf{x}^{1}_{i; 0}, \mathbf{x}^{1}_{i; K_{T}}, \mathbf{x}^{1}_{i; 2K_{T}}, ... , \mathbf{x}^{1}_{i; T})$, which is a sequence of $\frac{T}{K_{T}}$ frames $\mathbf{x}^{1}_{i; t} \in \mathbb{R}^{\frac{H}{K_{S}} \times \frac{W}{K_{S}} \times 3}$. 
The output of the first level is spatially upscaled and temporally interpolated by one or more subsequent upscaling levels. 

\textbf{Training}
We train our model greedily one level at a time and in order, i.e.\ we train the first level to generate global but downscaled videos, and then we train upscaling stages on previous level outputs one after each other.
We do not train the levels in an end-to-end fashion, which allows us to break down the computation into tractable steps by only training one level at a time.
We formulate a GAN objective for each stage of our model.
We consider the distribution $p_{g_1}$ in eq.~\ref{eq:gen_f} and solve a min-max game with the following value function:

{\small
\begin{eqnarray}
\mathbb{E}_{\mathbf{x}^1\sim p_{d}} [\log (D_1(\mathbf{x}^1))] + \mathbb{E}_{\mathbf{z}_1 \sim p_{z_1}} [\log (1-D_1(G_1(\mathbf{z}_1)))],
\label{eq:gan_stage1}
\end{eqnarray}} where $G_1$ and $D_1$ are the generator/discriminator associated with the first stage and $p_{z_1}$ is a noise distribution.
This is the standard GAN objective~\cite{goodfellow2014generative}. 
For upscaling levels corresponding to $p_{g_l}, l > 1$, we consider the following value function:

{\small
\begin{eqnarray}
&  \mathbb{E}_{\mathbf{x}^{l-1}, ..., \mathbf{x}^{1}\sim p_{d}} \mathbb{E}_{\mathbf{x}^l\sim p_{d}( .| \mathbf{x}^{l-1}, ..., \mathbf{x}^{1})} [\log(D_l(\mathbf{x}^{l}, \mathbf{x}^{l-1}))] + \nonumber \\ & \mathbb{E}_{\mathbf{\hat{x}}^{l-1}\sim p_{g_{l-1}}} \mathbb{E}_{\mathbf{z}_l\sim p_{z_l}} [\log(1-D_l(G_l(\mathbf{z}_l, \mathbf{\hat{x}}^{l-1}),  \mathbf{\hat{x}}^{l-1}))],
\label{eq:gan_stage2}
\end{eqnarray}} where $G_l$, $D_l$ are the generator and discriminator of the current level and $p_{g_{l-1}}$ is the generative distribution of the level $l-1$.
The min-max game associated with this value function has a global minimum when the two joint distributions are equal, $p_{d}(\mathbf{x}, ..., \mathbf{x^l}) = p_{g_l}(\mathbf{x^l} | \mathbf{x}^{l-1}) .. p_{g_1}(\mathbf{x}^1)$~\cite{dumoulin2016adversarially, donahue2016adversarial}.
We also see from eq.~\ref{eq:gan_stage2} that the discriminator for upscaling stages operates on pairs $(\mathbf{x}^l, \mathbf{x}^{l-1})$ videos to determine whether they are real or fake. 
This ensures that the upscaling stages are grounded on their inputs, i.e that $\mathbf{x}^l$ "matches" its corresponding $ \mathbf{x}^{l-1}$ .

\textbf{Partial View Training}
Computational requirements for upscaling levels can be high when generating large outputs. 
As we increase the length and resolution of a generation, the need to store activation tensors during training increases the amount of GPU memory required.
To further reduce the computational requirements, we propose to train the upscaling levels on only temporal crops of their inputs.
This strategy reduces training costs since we upscale smaller tensors, at the expense of having less available context to interpolate frames.
We define convolutional upscaling levels that learn functions that can be applied in a sliding window manner over their inputs.
At inference time, we do not crop the inputs and \ourmodel is applied to all possible input windows, thus generating full-length videos.

\section{Model Parametrization}
\label{section:model_parametrization}

In this section we describe the parametrization of the different levels of \ourmodel.
We keep the discussion at a high level, briefly mentioning the main components of our model.
Precise details on the architecture are provided in the appendix.
 
\textbf{First Level} 
The first level generator stacks units composed by a ConvGRU layer~\cite{ballas2015delving}, modeling temporal information, and 2D-ResNet blocks that upsample the spatial resolution.
Similar to MoCoGAN~\cite{tulyakov2018mocogan} and \dvdgan~\cite{clark2019efficient}, we use a dual discriminator with both a spatial discriminator that randomly samples $k$ full-resolution frames and discriminates them individually, and a temporal discriminator that processes spatially downsampled but full-length videos.

\textbf{Upsampling Levels} 
The upsampling levels are composed by a conditional generator and three discriminators (spatial, temporal and matching). 
The conditional generator produces an upscaled version $\mathbf{\hat{x}^l}$ of a lower resolution video $\mathbf{\hat{x}^{l-1}}$.
To discriminate samples from real videos, upscaling stages use a spatial and temporal discriminator, as in the first level.
Additionally, we introduce a matching discriminator.
The goal of the matching discriminator is to ensure that the output is a valid upsampling of the input, and its necessity arises from the model formulation.
Without this discriminator, the upsampling generator could learn to ignore the low resolution input video.
The conditional generator is trained jointly with the spatial, temporal and matching discriminators.

\textbf{Conditional Generator} 
The conditional generator takes as input a lower resolution video $\mathbf{\hat{x}}^{l-1}$, a noise vector $\mathbf{z}$ and optionally a class label $y$, and generates $\mathbf{\hat{x}}^{l}$.
Our conditional generator stacks units composed by one 3D-ResNet block and two 2D-ResNet blocks. 
Spatial upsampling is performed gradually by progressively increasing the resolution of the generator blocks.
To condition the generator we add residual connections~\cite{he2016deep, srivastava2015highway} from the low-resolution video to the output of each generator unit.

\textbf{Matching Discriminator} 
The matching discriminator uses an architecture like that of the temporal discriminator. It discriminates real or generated input-output pairs.
The output is downsampled to the same size as the input, and both tensors are concatenated on the channel dimension.
A precise description of all discriminator architectures can be found in the appendix.

\section{Related Work}

The modern video generation literature~\cite{ranzato2014video, video_lstm} first started as a result of adapting techniques for language modeling to video. 
Since then, many papers have proposed different approaches to represent and generate videos~\cite{luc2017predicting, luc2018predicting,villegas2017decomposing, villegas2017learning,xue2016visual}, including different kinds of tasks, conditionings and models.
We review the most common types of generative video models below.

Autoregressive models~\cite{larochelle2011neural,dinh2016density,kalchbrenner2017video,reed2017parallel,weissenborn2019scaling, yan2021videogpt} model the conditional probability of each pixel value given the previous ones.
They do not use latent variables and their training can be easily parallelized.
Inference in autoregressive models often requires a full forward pass for each output pixel, which does not scale well to long high resolution videos.
Normalizing flows~\cite{rezende2015variational,kingma2018glow,kumar2019videoflow} learn bijective functions that transform latent variables into data samples.
Normalizing flows are able to directly maximize the data likelihood. However, they require the latent variable to have the same dimensionality as its output, which becomes an obstacle when generating videos due to their large dimensionality.
Variational AutoEncoders (VAEs)~\cite{kingma2013auto, rezende2014stochastic, sv2p} also transform latent variables into data samples.
While more scalable, VAEs often produce blurry results when compared to other generative models.
Models based on VRNNs~\cite{vrnn, svg, savp, castrejon2019improved} use one latent variable per video frame and often produce better results.

Generative Adversarial Networks (GANs) are also latent variable models and optimize a min-max game between a generator G and a discriminator D trained to tell real and generated data apart~\cite{goodfellow2014generative}.
Empirically, GANs usually produce better samples than competing approaches but might suffer from mode collapse.
GAN models for video were first proposed in~\cite{vondrick2016generating, vondrick2016anticipating, mathieu2015deep}.
In recent work, SAVP~\cite{savp} proposed to use the VAE-GAN~\cite{larsen2015autoencoding} framework for video. 
TGANv2~\cite{saito2018tganv2} improves upon TGAN~\cite{TGAN2017} and proposes a video GAN trained on data windows, similar to our approach. 
However, unlike TGANv2, our model is composed of multiple stages which are not trained jointly.
MoCoGAN~\cite{tulyakov2018mocogan} first introduced a dual discriminator architecture for video, with \dvdgan~\cite{clark2019efficient} scaling up this approach to high resolution videos in the wild.
\dvdgan outperforms MoCoGAN and TGANv2, and is arguably the current state-of-the-art in adversarial video generation.
Our model is also related to work that proposes hierarchical or progressive training approaches for generative models~\cite{karras2017progressive,denton2015deep,xiong2018learning,zhao2020towards}. 
Our model is different in that our stages are trained greedily in separate steps without backpropagation from one stage to the other, which reduces its computational requirements. 

\section{Experiments}

\begin{figure*}[!ht]
    \centering
    \includegraphics[trim={0 78 0 0}, clip, width=0.85\textwidth]{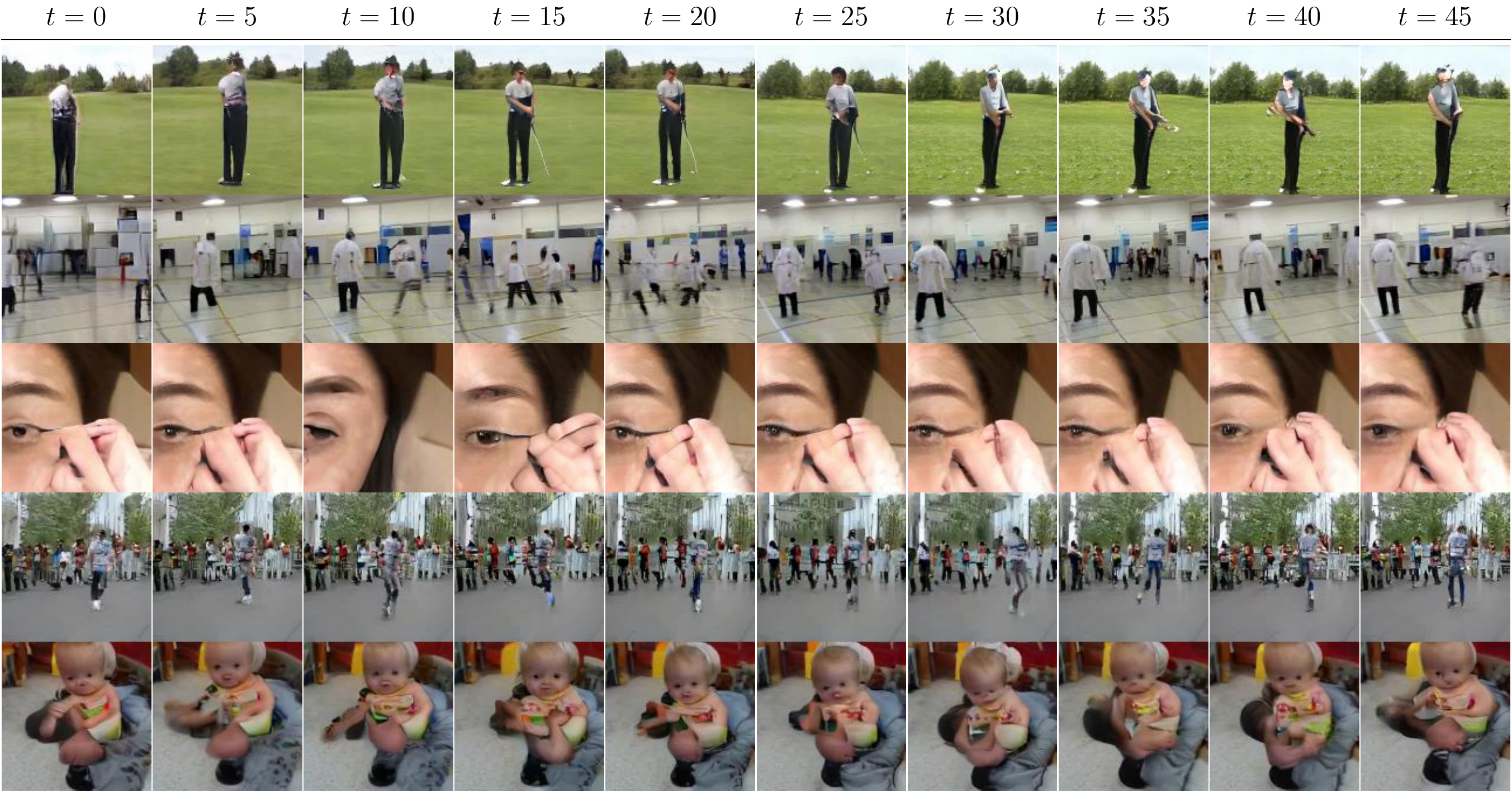}
    \caption{
    \small \textbf{Randomly selected \ourmodel 48/128x128 frame samples for Kinetics-600:} 
    These samples were generated by unrolling \ourmodel 12/128x128 to generate 48 frame sequences, 4 times its training horizon. 
    Each row shows frames from the same sample at different timesteps.
    The generations are temporally consistent and the frame quality does not degrade over time.
    }
    \label{fig:samples_kinetics_128}
\end{figure*}

\begin{table*}
    \caption{\small \textbf{Results on Kinetics-600 128x128} We compare our two-level \ourmodel against the reported metrics for \dvdgan ~\cite{clark2019efficient}. Our model is trained on 12-frame windows and matches the performance of the 12-frame \dvdgan model. Furthermore, the same \ourmodel model is able to generate 48 frames when applied convolutionally over a full-length first level output.
    In that setup our model also matches the quality of a 48-frame \dvdgan model, but has significantly lower computational requirements.}
    \label{tab:dvdgan_comparison}
    \centering
    \begin{tabular}{lccccccc}
    \toprule
    & & \multicolumn{3}{c}{Evaluated on 12 frames} & \multicolumn{3}{c}{Evaluated on 48 frames}\\
    \cmidrule(lr){3-5} \cmidrule(lr){6-8} 
    Model & Trained on & IS ($\uparrow$) & FID ($\downarrow$) & FVD ($\downarrow$) & IS ($\uparrow$) & FID ($\downarrow$)  & FVD ($\downarrow$) \\
    \midrule
          DVD-GAN & 12/128x128 & 77.45 & \textbf{1.16} & - & N/A & N/A & N/A \\
          DVD-GAN & 48/128x128 & N/A & N/A & N/A & \textbf{81.41} & 28.44 & -  \\
    \midrule
          2-Level \ourmodel & 12/128x128 & \textbf{104.00} & 2.09 & 591.90 & 77.36 & \textbf{14.00} & 517.21 \\
    \bottomrule
    \end{tabular}
    
\end{table*}

\begin{figure}
    \centering
    \includegraphics[trim={10 10 10 10},clip,width=0.85\linewidth]{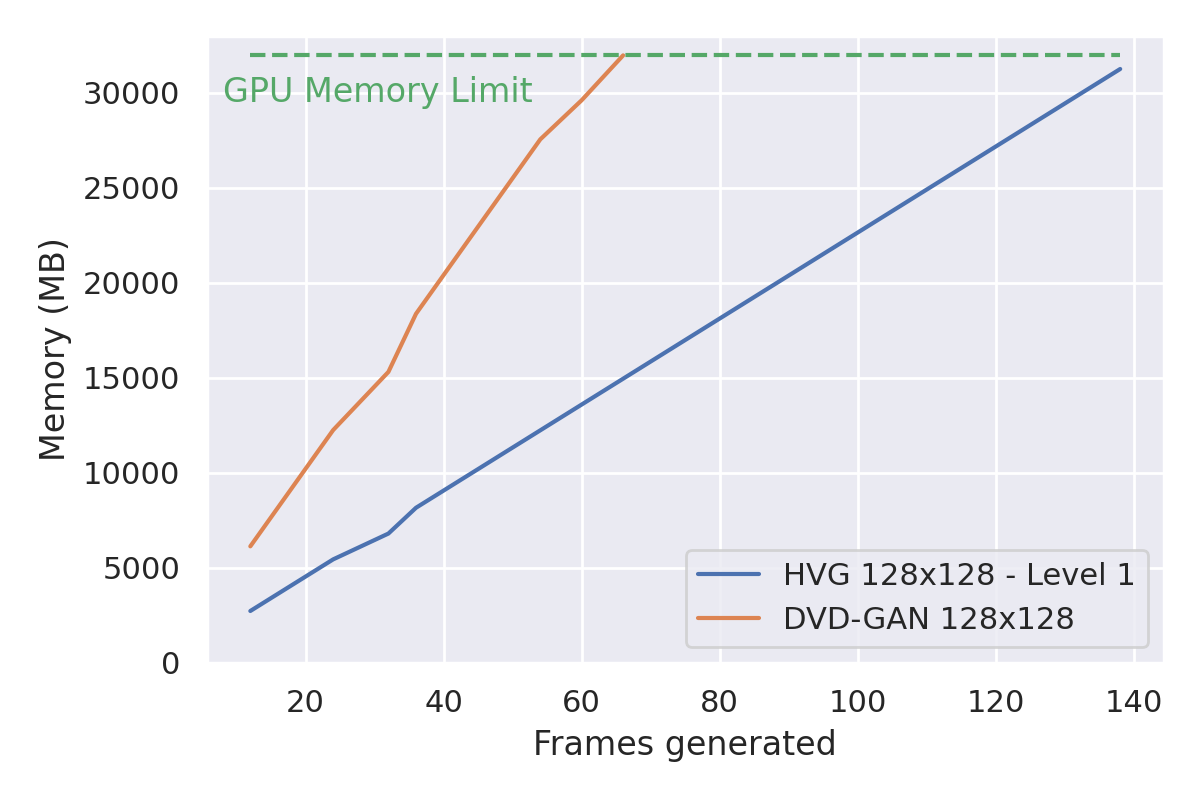}
    \caption{\small \textbf{Scaling the computational costs}
    We report the required GPU memory for a two-level \ourmodel. We observe that, given the same batch size, the memory cost scales linearly with the output length. Our model scales better than a comparable non-cascaded model.}
    \label{fig:computational_costs}
\end{figure}

\begin{figure*}
    \centering
    \includegraphics[trim={0 100 0 0}, clip, width=0.85\textwidth]{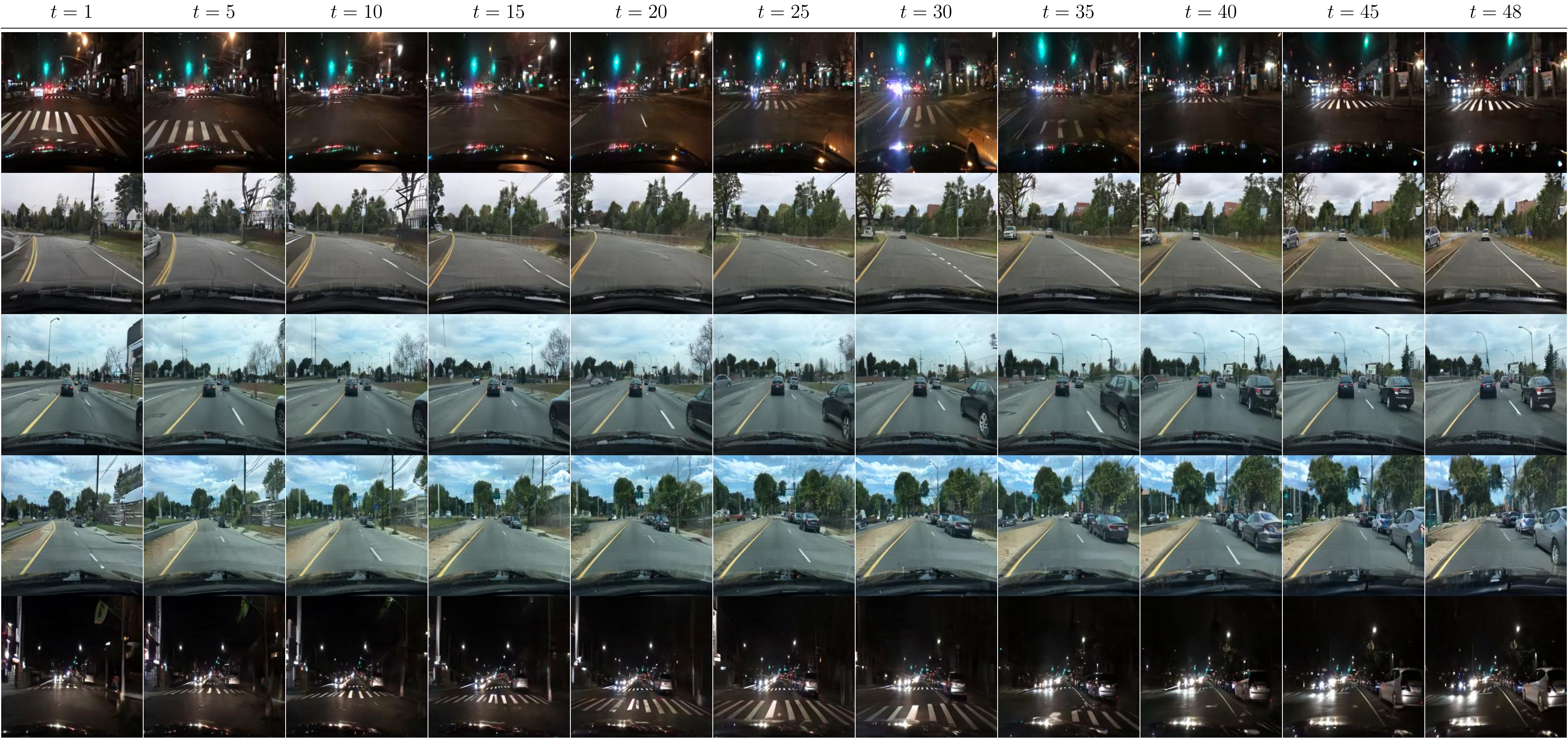}
    \caption{\small \textbf{Random 48/256x256 BDD100K samples:} We show samples from our three-stage BDD100K model. 
    Each row shows a different sample over time. 
    Despite the two stages of local upsampling, the frame quality does not degrade noticeably through time. 
    }
    \label{fig:samples_bdd_256}
\end{figure*}

In this section we empirically validate our proposed approach.
First, we show that our approach outperforms or matches the state-of-the-art on Kinetics-600 and UCF101.
Then, we analyze the scaling properties of our model in Section~\ref{sec:bdd100k}.
Finally, we ablate the main components of our model in Section~\ref{sec:ablations}.

\subsection{Experimental Setting}
\textbf{Datasets}
We consider the Kinetics-600~\cite{kay2017kinetics,carreira2018short} and the UCF101~\cite{soomro2012ucf101} datasets for class conditional video generation. 
Kinetics-600 is a large scale dataset of Youtube videos depicting 600 action classes.
The videos are captured in the wild and exhibit lots of variability. 
The amount of videos available from Kinetics-600 is constantly changing as videos become unavailable from the platform.
We use a version of the dataset collected on June 2018 with around 350K videos.
UCF101 contains approximately 13K videos with around 27 hours of video from 101 human action categories. Its videos have camera motion and cluttered backgrounds, and it is a common benchmark in the video generation community.

Additionally, we use the BDD100K dataset~\cite{yu2018bdd100k} for unconditional video generation. 
BDD100K contains 100k videos showing more than 1000 hours of driving under different conditions. 
We use the training set split of 70K videos.

For the rest of the section we denote video dimensions by their output resolution DxD and number of frames F as F/DxD.

\textbf{Evaluation metrics}
Defining evaluation metrics for video generation is an open research area. 
We use metrics from the image generation literature adapted to video.
On Kinetics, we report three metrics: i) Inception Score (IS) given by an I3D model~\cite{carreira2017quo} trained on Kinetics-400, ii) Frechet Inception Distance on logits from the same I3D network, also known as Frechet Video Distance (FVD)~\cite{fvd}, and iii) Frechet Inception Distance on the last layer activations of an I3D network trained on Kinetics-600 (FID).
On BDD100K we report FVD and FID as described before, but we omit IS scores as they are not applicable since there are no classes.
On UCF101 we report IS scores following the standard setup in the literature.

\textbf{Implementation details}
All \ourmodel models are trained with a batch size of 512 and using up to 4 nVidia DGX-1.
Levels for Kinetics-600 are trained for 300k iterations, while levels for BDD100K and UCF101 are trained for 100k iterations, all with early stopping when evaluation metrics stop improving.
We use PyTorch and distribute training across multiple machines using data parallelism. 
We synchronize the batch norm statistics across workers.
We employ the Adam~\cite{kingma2014adam} optimizer to train all levels with a fixed learning rate of \num{1e-4} for G and \num{5e-4} for D.
We use orthogonal initializations for all weights in our model and spectral norm in the generator and the discriminator.
More details can be found in the appendix. 

\textbf{Baselines}
As baselines we consider \dvdgan~\cite{clark2019efficient}, TGANv2~\cite{TGAN2017, saito2018tganv2}, VideoGPT~\cite{yan2021videogpt} and MoCoGAN~\cite{tulyakov2018mocogan}. 
Comparisons are mostly against DVD-GAN, as the current state-of-the-art model for class-conditional video generation and the only approach that can generate realistic samples on Kinetics-600.

\subsection{Kinetics-600}
\label{sec:kinetics600}

We first evaluate the performance of \ourmodel on the Kinetics-600 dataset with complex natural videos.

We train a two-level \ourmodel on Kinetics-600 that generates either 12/128x128 or 48/128x128 videos, to compare to the previous state-of-the-art.
The first level of \ourmodel generates 24/32x32 videos with a temporal subsampling of 4 frames.
The second level upsamples the first level output using a factor of 2 for the temporal resolution and a factor 4 for the spatial resolution, producing 48/128x128 videos with a temporal subsampling of 2 frames.
Since these generations are large, we employ partial views on first level outputs to train the second level, which takes as input windows of 6/32x32 frames and is trained to generate 12/128x128 video snippets (4x lower dimensional than the final output).
As a result, this level has approximately the same training cost than a model generating videos of size 12/128x128.
At inference time we run the second level convolutionally over all the 24 first level frames to generate 128x128 videos with 48 frames. 
We also use the same model to generate 12/128x128 videos by using random 6 frame windows from the first level output (i.e. we use the same training and inference setup).
We compare to \dvdgan for both 12/128x128 and 48/128x128 videos with temporal subsampling 2, as the current state-of-the-art in this dataset. Note that \dvdgan outperforms by a large margin other approaches on Kinetics-600 such as~\cite{weissenborn2019scaling}.

We report the scores obtained by our model in Table~\ref{tab:dvdgan_comparison}. 
For 12/128x128 videos, our model achieves higher IS and comparable FID to \dvdgan, validating that both models perform comparably when using a similar amount of computational resources.
Additionally, \ourmodel outperforms a 48/128x128 DVD-GAN model in FID score and reaches a similar IS score, despite only being trained on reduced views of the data.
Qualitatively, the generations of both models are similar - they do not degrade noticeable in appearance through time although both have some temporal inconsistencies. 
For \ourmodel temporal inconsistencies are often due to a poor first level generation.
For \dvdgan we hypothesize that inconsistencies are due to the reduced temporal field of view of its discriminator, which is of less than 10 frames while the model has to generate 48 frames.

\subsection{UCF101}
\label{sec:ucf101}

We further evaluate our approach on the smaller UCF101 dataset.
We train a two-level \ourmodel to generate 16/128x128 videos, as commonly done in the literature.
The first level generates 8/64x64 videos with a temporal subsampling of 2 frames. 
The second level upscales the output of the first level to 16/128x128 videos, without temporal subsampling to match the literature.
Since these are small videos, we do not use temporal windows to train the second upscaling level.

We report the scores obtained by our model in Table~\ref{tab:ucf101_comparison}. 
Our model obtains state-of-the-art results, outperforming previous approaches by a large margin.
Qualitatively, \ourmodel generates coherent samples with high fidelity details, with small temporal inconsistencies.
Samples for the UCF101 dataset can be found in the appendix.

\begin{table}[t]
    \caption{\small \textbf{UCF101 16/128x128 comparison:} We compare \ourmodel to previous video generation approaches on UCF101. Our method obtains significantly higher Inception Score.}
    \label{tab:ucf101_comparison}
    \centering
    \begin{tabular}{c|c}
        \toprule
        \textbf{Model} & \textbf{IS($\uparrow$)} \\
        \midrule
        MoCoGAN~\cite{tulyakov2018mocogan} & 12.42 \\
        VideoGPT~\cite{yan2021videogpt} & 24.69 \\
        TGANv2~\cite{saito2018tganv2} & 28.87 \\
        DVD-GAN~\cite{clark2019efficient} & 32.97 \\
        \midrule
        \ourmodel (ours) & \textbf{53.72} \\
        \bottomrule
    \end{tabular}
\end{table}

\begin{figure*}[t]
    \centering
    \includegraphics[trim={0 0 0 0},clip,width=0.85\textwidth]{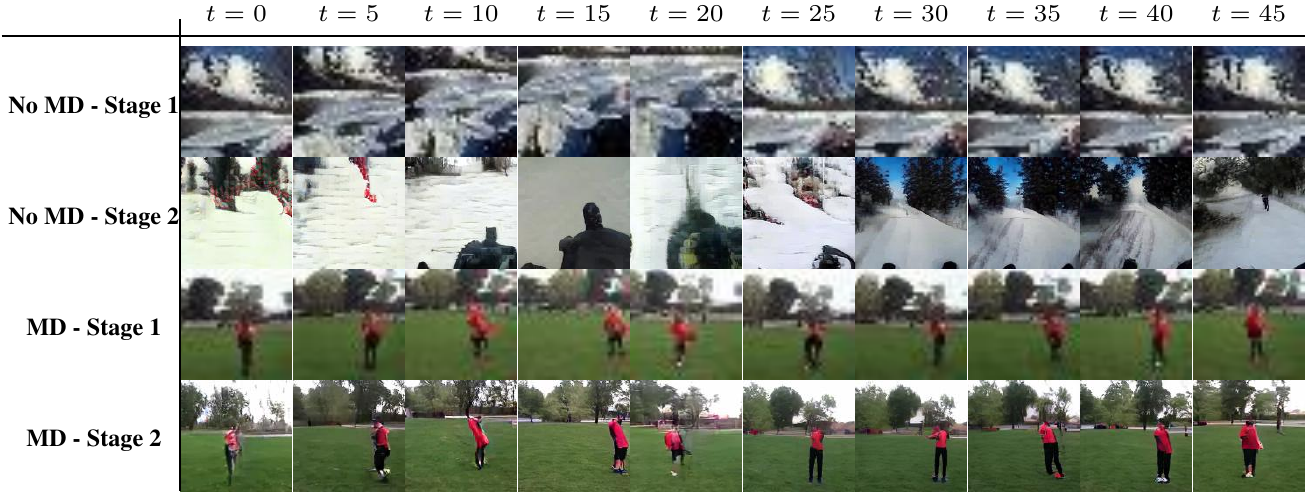}
    \caption{\small \textbf{Matching discriminator samples} We show a random sample from our two-level model on Kinetics-600 with the matching discriminator and without the matching discriminator (No MD). For each sample we show the output of the first level and the corresponding second level output. While the No MD model generates plausible local snippets at level 2, it does not remain temporally coherent. Our model with the matching discriminator is temporally consistent because it is grounded in the low resolution input.}
    \label{fig:md_samples}
\end{figure*}

\subsection{Scaling up \ourmodel on BDD100K}
\label{sec:bdd100k}
To show that our model scales well with the video dimensionality, we train a three-stage model on the BDD dataset to generate 48/256x256 videos.
A training iteration with a single 48/256x256 video for a similarly sized non-cascaded model requires more than 32GB of GPU memory.
Such model is therefore not trainable on most current GPUs without techniques like gradient checkpointing which add a significant overhead to the training time.
Instead, with our cascaded model we can fit 4 examples per GPU without any engineering tricks.

We train the first level to output 12 frames at 64x64 resolution with a temporal subsampling of 8 frames.
The second level upsamples 12 frame windows at 128x128 resolution with temporal subsampling of 4 frames (since we are doubling the framerate of the first level).
The third level is trained to upscale 12 frame windows at 256x256 resolution for a final temporal subsampling of 2 frames.
Figure~\ref{fig:samples_bdd_256} shows samples from this model.
The videos appear crisp, show multiple settings and do not degrade through time.

To further illustrate the scaling capabilities of our model, we report the memory requirements for a two-level 128x128 \ourmodel as a function of the number of output frames in Figure~\ref{fig:computational_costs}.
The first level generates half the output frames at 64x64, while the second level is trained to upscale windows of 6 frames into 12 128x128 frames, regardless of the first level output length.
Compared to a non-cascaded 128x128 model, our first level scales better due to the lower resolution and reduced number of frames, and the second level has a fixed memory cost of 10290MB since it is trained on 6 frames windows.
Given the same GPU memory budget, our model can generate sequences of up to 140 frames, more than double the frames of a non-cascaded model.

\subsection{Model Ablations}
\label{sec:ablations}

\begingroup
\def\arraystretch{0.9}
\begin{table*}
\caption{\small \textbf{Matching discriminator comparison} 
We report metrics on Kinetics-600 and BDD100K for our model with and without the matching discriminator. 
Both models perform similarly for 6 frames, corresponding to the training video length. 
However, the model without the matching discriminator produces incoherent generations when applied over the full first level input because it can ignore it.
}
\label{tab:results_kinetics}
\center
\begin{tabular}{cccccccc}
\toprule
& & \multicolumn{3}{c}{6 Frames} & \multicolumn{3}{c}{50 Frames} \\
\cmidrule(lr){3-5} \cmidrule(lr){6-8}
Dataset & Model & IS ($\uparrow$) & FID ($\downarrow$) & FVD ($\downarrow$) & IS ($\uparrow$) & FID ($\downarrow$) & FVD ($\downarrow$) \\
\midrule
  \multirow{2}{*}{Kinetics-600}  
& \ourmodel (No MD) & \textbf{50.31} & 1.62 & 594.99 & 37.81 & 42.29 & 1037.79 \\
& \ourmodel & 48.44 & \textbf{1.06} & \textbf{565.95} & \textbf{49.44} & \textbf{31.87} & \textbf{790.97} \\
\midrule
\multirow{2}{*}{BDD100K} & \ourmodel (No MD) & N/A & 1.36 & 211.69 & N/A & 26.52 & 575.51 \\
& \ourmodel & N/A & \textbf{1.07} & \textbf{144.96} & N/A & \textbf{18.73} & \textbf{326.78} \\
\bottomrule
\end{tabular}

\end{table*}
\endgroup

\textbf{Matching Discriminator Ablation}
To assess the importance of the matching discriminator, we compare two-level \ourmodel models with and without the matching discriminator (we refer to the latter as No MD). 
As in Section~\ref{sec:kinetics600}, we generate 48 frames with a two-level model on Kinetics-600.
We train the upscaling level on 3-frame windows of the first level to generate 6 frames.
We expect the No MD model to generate inconsistent full-length videos when applied over the full first level since its outputs are not necessarily valid upscalings of the inputs.
On 6 frame generations (i.e.\ the training setup), \ourmodel and \ourmodel No MD obtain similar scores as reported in Table~\ref{tab:results_kinetics}. 
While the No MD model ignores its previous level inputs, it still learns to generate plausible 6 frame videos at 128x128.
However, when we use the models to generate full-length 48 frame videos, \ourmodel No MD only generates valid local snippets and is inconsistent through time.
Fig.~\ref{fig:md_samples} shows an example of a full length No MD generation in which this effect is observable.
In contrast, our model (MD) stays grounded to the input and remains consistent through time.
This is reflected in the reported metrics in Table~\ref{tab:results_kinetics}, where the No MD model has worse scores. 
This ablation shows the need of a matching discriminator to ground upsampling level outputs to their inputs.

\begingroup
\begin{table}
\caption{\small \textbf{Temporal Window Ablation} 
We compare two-level models trained with different window sizes for the upscaling levels.
The 6-frame model has higher computational requirements but outperforms the 3-frame model, confirming a trade-off between computational savings and final performance when selecting the temporal window size.
}
\label{tab:window_ablation}
\center
\begin{tabular}{cccc}
\toprule
& \multicolumn{3}{c}{Kinetics-600 48 Frames} \\
\cmidrule(lr){2-4}
Window & IS ($\uparrow$) & FID ($\downarrow$) & FVD ($\downarrow$) \\
\midrule
3-frame  & 58.21 & 31.59 & 714.74 \\
6-frame  & \textbf{77.36} & \textbf{14.00} & \textbf{517.21} \\
\bottomrule
\end{tabular}

\end{table}
\endgroup

\textbf{Temporal Window Ablation}
One modelling choice in \ourmodel is the temporal window length used in the upsampling levels.
Shorter inputs provide less context to upsample frames, while longer inputs require more compute.
To assess the impact of the window length, we compare two-level models trained on Kinetics-600 128x128: one trained on first level windows of 6 frames (same setup as in Section~\ref{sec:kinetics600}) and one trained on windows of only 3 frames.
The 6-frame level requires approximately 2x GPU memory than the 3-frame level during training, but we expect it to perform better due to the larger context available for upscaling.
We compare their performance to generate 48 frames in Table~\ref{tab:window_ablation}.
We conclude that the window size defines a trade-off between computational resources and sample quality. 
We refer the reader to the appendix for an additional ablations and experiments.
\section{Conclusions}

We propose \ourmodel, a cascaded video generator that divides the generative process into simpler steps.
Our model is competitive with state-of-the-art approaches in terms of sample quality, while requiring significantly less computational resources due to the cascaded approach.
Higher capacity models and larger outputs are key aspects in improving video generation, and \ourmodel is a step in that direction with better scaling properties than previous approaches.

\subsection*{Acknowledgements}
We thank the Mila Quebec AI Institute for managing the computer clusters on which this research was conducted. 
This work was supported by an IVADO PhD Fellowship to L.C. and funding from CIFAR.

{\small
\bibliographystyle{IEEETran}
\bibliography{bibliography.bib}
}

\end{document}